\documentclass{article}

\usepackage{microtype}
\usepackage{graphicx}
\usepackage{subfigure}
\usepackage{booktabs} 
\usepackage{hyperref}
\usepackage{multirow}
\usepackage{enumitem}
\usepackage{graphics}
\usepackage{subfigure}
\usepackage{float}
\usepackage{diagbox}
\usepackage{array}
\usepackage{ragged2e}
\usepackage{pifont}
\usepackage{amsfonts}
\usepackage{graphicx}
\usepackage{caption}
\usepackage{subcaption}
\usepackage{bm}
\usepackage{graphicx} 
\usepackage{xspace}

\usepackage[accepted]{icml2024}

\usepackage{amsmath}

\usepackage{amssymb}
\usepackage{mathtools}
\usepackage{amsthm}
\usepackage{graphicx}
\usepackage{amsthm}
\usepackage{multirow}
\usepackage[capitalize,noabbrev]{cleveref}
\theoremstyle{plain}
\newtheorem{theorem}{Theorem}[section]

\theoremstyle{definition}

\newtheorem{problem}{Problem}
\newtheorem{assumption}[theorem]{Assumption}
\theoremstyle{remark}

\newcommand{\ourmethod}{FOIL\xspace}
\newcommand{\supportE}{\mathrm{supp}(\bm{E})\xspace}

\newcommand{\XI}{\bm{X}_{\mathrm{I}}}
\newcommand{\XV}{\bm{X}_{\mathrm{V}}}

\newcommand{\YSUF}{\bm{Y}^{\text{suf}}}
\newcommand{\CLD}{\bm{\mathcal{C}}_{\mathrm{LD}}}
\newcommand{\MTEI}{\bm{\mathcal{M}}_{\mathrm{TEI}}}
\newcommand{\MTIL}{\bm{\mathcal{M}}_{\mathrm{TIL}}}
\newcommand{\LSUF}{\ell_{\text{suf}}}

\icmltitlerunning{Time-Series Forecasting for Out-of-Distribution Generalization Using Invariant Learning}
\usepackage{xcolor} 
\begin{document}

\twocolumn[
\icmltitle{\textbf{Time-Series Forecasting for Out-of-Distribution Generalization\\ Using Invariant Learning}}


\icmlsetsymbol{equal}{*}

\begin{icmlauthorlist}
\icmlauthor{Haoxin Liu}{yyy}
\icmlauthor{Harshavardhan Kamarthi}{yyy}
\icmlauthor{Lingkai Kong}{yyy}
\icmlauthor{Zhiyuan Zhao}{yyy}
\icmlauthor{Chao Zhang}{yyy}
\icmlauthor{B. Aditya Prakash}{yyy}

\end{icmlauthorlist}

\icmlaffiliation{yyy}{School of Computational Science and Engineering, Georgia
Institute of Technology, Atlanta, USA}

\icmlcorrespondingauthor{Haoxin Liu}{hliu763@gatech.edu}
\icmlcorrespondingauthor{B. Aditya Prakash}{badityap@cc.gatech.edu}

\icmlkeywords{Machine Learning, ICML}

\vskip 0.3in
]



   

    





\printAffiliationsAndNotice{}
\begin{abstract}
Time-series forecasting (TSF) finds broad applications in real-world scenarios. Due to the dynamic nature of time-series data, it is crucial to equip TSF models with out-of-distribution (OOD) generalization abilities, as historical training data and future test data can have different distributions.
In this paper, we aim to alleviate the inherent OOD problem in TSF via invariant learning. We identify fundamental challenges of invariant learning for TSF. 
First, the target variables in TSF may not be sufficiently determined by the input due to unobserved core variables in TSF, breaking the conventional assumption of invariant learning. 
Second, time-series datasets lack adequate environment labels, while existing environmental inference methods are not suitable for TSF. 

To address these challenges, we propose \ourmethod, a model-agnostic framework that enables time-series \underline{\textbf{F}}orecasting for \underline{\textbf{O}}ut-of-distribution generalization via \underline{\textbf{I}}nvariant \underline{\textbf{L}}earning. \ourmethod employs a novel surrogate loss to mitigate the impact of unobserved variables. Further, \ourmethod implements a joint optimization by alternately inferring environments effectively with a multi-head network while preserving the temporal adjacency structure, and learning invariant representations across inferred environments for OOD generalized TSF. We demonstrate that the proposed \ourmethod significantly improves the performance of various TSF models, achieving gains of up to 85\%.
\end{abstract}
\section{Introduction}\label{sec:intro}
Time-series (TS) data are ubiquitous across various domains, including public health~\cite{pub1,pub2}, finance~\cite{finance}, and urban computing~\cite{urban}. Time-series forecasting (TSF), a foundational task in analyzing TS data, involving predicting future events or trends based on historical TS data, has received a longstanding research focus. TSF faces certain challenges due to the dynamic and complex nature of TS data: First, distributions of TS data change over time. Second, the inherent complexity of TSF is compounded by unforeseen exogenous factors, such as policy interventions and climate changes in the context of influenza forecasting.

Given the dynamic nature of TS data, where unforeseen distribution shifts can occur between historical training and future testing data, the TSF task asks for robust out-of-distribution (OOD) generalization abilities. Instead, existing TSF models employ empirical risk minimization to greedily incorporate all correlations within the data to minimize average training errors. However, as not all correlations persist in unknown test distributions, these models may lack OOD generalization abilities. Note that existing works on temporal distribution shifts~\cite{adarnn,revin,Nonstationary,dish} merely focus on mitigating the marginal distribution shifts of the input. These methods are not generalizable enough for the OOD problem, which consists of various types of distribution shifts~\cite{OOD}, such as conditional distribution shifts, etc.

In this paper, we propose to alleviate the OOD generalization problem of TSF via invariant learning (IL). IL seeks to identify and utilize invariant features that maintain stable relationships with targets across different environments while discarding unstable correlations introduced by variant features. 
Although IL has witnessed wide theoretical and empirical success in various domains~\cite{th1,th2,th3}, it remains unexplored yet non-trivial to apply IL for TSF because of the following challenges: First, TS data breaks IL's conventional assumption.
In TS data, there are always variables that directly affect targets but remain unobserved, such as the outbreak of an epidemic, sudden temperature changes, policy adjustments, etc.
IL fails to consider these unobserved core variables, leading to poor OOD generalization in TSF.
Second, TS data are usually collected without explicit environment labels.
Although some general IL with environment inference methods have been proposed, their neglect of TS data characteristics results in suboptimal inferred time-series environments. 

Thus, we propose a novel TSF approach for out-of-distribution generalization, namely \ourmethod (\underline{F}orecasting for \underline{O}ut-of-distribution TS generalization via \underline{I}nvariant \underline{L}earning). Our contributions are summarized as follows:
\begin{itemize}[noitemsep]
    \item We investigate the out-of-distribution generalization problem of time-series forecasting. To the best of our knowledge, we are the first to introduce invariant learning to TSF and identify two essential gaps, including the non-compliance of IL's conventional assumption and the lack of environment labels.

    \item We propose \ourmethod, a practical and model-agnostic invariant learning framework for TSF. \ourmethod leverages a simple surrogate loss to ensure the applicability of IL and designs an efficient environment inference module tailored for time-series data. 
 
    \item We conduct extensive experiments on diverse datasets along with three advanced forecasting models (`backbones'). \ourmethod proves effectiveness by uniformly outperforming all baselines in better forecasting accuracy.

\end{itemize}

\section{Preliminaries and Problem Definition}\label{sec:definition}
We formally introduce the TSF task and discuss why it is an OOD generalization problem. We then introduce the problem OOD-TSF, formulating TSF as an OOD problem.

We denote slanted upper-cased letters such as \( \bm{X}\) as random variables and calligraphic font letters \( \mathcal{X} \) as its sample space. Upright bold upper-cased letters such as \( \mathbf{X}  \), bold lower-cased letters such as \( \bm{x} \) and regular lower-cased letters such as \( x \) denote deterministic matrices, vectors and scalars, respectively.
\subsection{Time-Series Forecasting: An Out-of-Distribution Generalization View} 

TSF models take a time series as input and output future values of some or all of its features. Let the input time-series variable be denoted as $\bm{X}\in\mathbb{R}^{l\times d_\mathrm{in}}$, where $l$ is the length of the \textit{lookback window} decided by domain experts and $d_\mathrm{in}$ is the feature dimension at each time step. 
The output variable of the forecasts generated of \textit{horizon window} length $h$ is denoted as $\bm{Y}\in\mathbb{R}^{h\times d_\mathrm{out}}$, where $d_\mathrm{out}$ is the dimension of targets at each time step. For the sample at time step $t$, denoted as $(\mathbf{X}_t,\mathbf{Y}_t)$, $\mathbf{X}_{t}\in\bm{X}=\left[\mathbf{x}_{t-l+1}, \mathbf{x}_{t-l+2}, \ldots, \mathbf{x}_{t}\right]$ and $\mathbf{Y}_{t}\in\bm{Y}=\left[\mathbf{y}_{t+1}, \mathbf{y}_{t+2}, \ldots, \mathbf{y}_{t+h}\right]$. Thus, the TSF model parameterized by $\theta$ is denoted as $f_\theta:\mathcal{X}\rightarrow\mathcal{Y}$. 

In this paper, we focus on univariate forecasting with covariates, i.e., \( d_{\text{out}} = 1 \) and \( d_{\text{in}} \geq 1 \), but our method can be easily generalized to the multivariate forecasting setting by using multiple univariate forecasting~\cite{M1,M2}. 

Existing TSF models usually assume the training distribution is the same as the test distribution and use empirical risk minimization (ERM) for model training.
However, training and test sets of TSF represent historical and future data, respectively. 
Given the dynamic nature of time series, the test distribution may diverge from the training distribution. 
In this paper, we consider TSF under the more realistic situation where  \( P^{\mathrm{train}}(\boldsymbol{X}, \boldsymbol{Y}) \neq P^{\mathrm{test}}(\boldsymbol{X}, \boldsymbol{Y}) \), 
i.e., unknown \( P^{\mathrm{test}}(\boldsymbol{X}, \boldsymbol{Y}) \), which can be  defined as follows:

\begin{problem} Out-of-Distribution Generalization for Time-Series Forecasting (\textbf{OOD-TSF}):
  Given a time-series training dataset $\mathcal
{D}^{\mathrm{train}}=\{(\mathbf{X}_{t},\mathbf{Y}_{t})\}_{t=1}^{T}$, the task is to learn an out-of-distribution generalized forecasting model $f_{\theta}^{*}:\mathcal{X}\rightarrow\mathcal{Y}$ parameterized by $\theta$ which achieves minimum error on testing set $\mathbf{\mathcal{D}}^{\mathrm{test}}$ with unknown distribution $P^{\mathrm{test}}(\bm{X},\bm{Y})$.
\end{problem}\label{problem}

\subsection{Invariant Learning: Out-of-Distribution Generalization with Environments}

\par \noindent \textbf{Environment Labels}. Invariant learning (IL), backed by the invariance principle~\cite{IRM} from causality, is a popular solution for OOD generalization. IL assumes heterogeneity in observed data: dataset is collected from multiple environments, formulated as \( \mathbf{\mathcal{D}} = \cup_{e} \mathbf{\mathcal{D}}^e  =\cup_{e}\{(\mathbf{X}^{e}_{i},\mathbf{Y}^{e}_{i})\}_{i=1}^{|\mathbf{\mathcal{D}}^e|}\); each environment $e$ has a distinct distribution $P^{e}(\boldsymbol{X},\boldsymbol{Y})$, termed heterogeneous environments. In time-series data, temporal environments can be seasons, temperatures, policies, etc. Let $\supportE$ denote all environments, the objective function is formulated as:
 \begin{equation}
    \bm{\mathcal{R}}_{\mathrm{IL}}(f_{\theta})=\max_{e\in \mathrm{supp}(\bm{E})}\mathbb{E}_{P(\bm{X},\bm{Y}|e)}\left[\ell(f_{\theta}(\mathbf{X}),\mathbf{Y}))|e\right],
\label{obj:IL}
\end{equation}
where OOD generalization is achieved by minimizing the empirical risk under the worst-performing environment. 
\par \noindent \textbf{Invariant Features.} To optimize Eq.~\ref{obj:IL}, IL proposes to identify and utilize invariant features that maintain stable relationships with target variables across different environments. 
For instance, in forecasting the number of flu cases, temperature changes belong to invariant features~\cite{temp1,temp2}, while hospital records are variant features since the proportion of influenza cases over all records may vary across different seasons.

\par \noindent \textbf{Sufficiency and Invariance Assumption.}
Most IL methods are proposed based on the following conventional assumption~\cite{as2,as1,InvariantRationalization,IRM,HRM,ZIN}:
\begin{assumption}[Conventional Assumption of Invariant Learning]
The input features $\bm{X}$ is a mixture of invariant features $\XI$ and variant features $\XV$. $\XI$ possesses the following properties:
\begin{itemize}
\setlength{\itemsep}{0pt} 
    \item[\textbf{a.}] \textbf{Sufficiency property:} \( \bm{Y}= g(\XI) + \epsilon, \) where  \( g(\cdot)\) can be any mapping function, and $\epsilon$ is random noise.
    \item[\textbf{b.}] \textbf{Invariance property:} for all $ e_{i}, e_{j} \in \mathrm{supp}(\bm{E})$, we have $P^{e_{i}}(\bm{Y} | \XI) = P^{e_{j}}(\bm{Y} | \XI)$ holds.
\end{itemize}
\label{assump:fund}
\end{assumption}
Thus, $\XI$ is assumed to provide sufficient and invariant predictive power for $\bm{Y}$ and is theoretically proven to guarantee optimal OOD performance for Eq.~\ref{obj:IL}~\cite{HRM}.

To better understand the above, we employ the structural causal model (SCM)~\cite{SCM} shown in Figure~\ref{fig:CG_A}. We define invariant features $\XI$ as the subset of input features $\bm{X}$ that directly cause $\bm{Y}$, following ~\cite{IRM,causal,ZIN}. Environment $\bm{E}$ can be interpreted as the confounder between \(\XI\) and \(\XV\). Specifically, the correlation between \(\XV\) and \(\bm{Y}\) is spurious, mediated through \(\XV \leftarrow \bm{E} \rightarrow \XI \rightarrow \bm{Y}\). Conversely, the causal relationship \(\XI \rightarrow \bm{Y}\) is invariant. Generally, IL aims to achieve OOD generalization using such $\XI$ to predict $\bm{Y}$.

\section{Challenges}\label{sec:gap}
Considering the theoretical and empirical successes of invariant learning~\cite{IRM,th1,VREx,th2}, a natural question arises: \textbf{Can we directly apply invariant learning (IL) to OOD-TSF?} Unfortunately, there are two main reasons rendering a direct application problematic. Firstly, the existence of unobserved variables in time-series (TS) data breaks the conventional Assumption~\ref{assump:fund} of IL. Secondly, TS datasets usually lack adequate environment labels. 
\par \noindent \textbf{TS data break IL's conventional assumption.} 
Recall Assumption~\ref{assump:fund}, where invariant features $\XI$ are assumed to provide sufficient and invariant predictive power for $\bm{Y}$ in IL. However, in TSF tasks, there are always variables that directly affect $\bm{Y}$ but are not included in the input features $\bm{X}$, such as the outbreak of a novel epidemic, sudden temperature changes, policy adjustments, etc. These \textit{unobserved core variables}, denoted as $\bm{Z}$, exist due to their absence from the whole\ dataset or the lookback window.

\begin{figure}[h]
\centering
\subfigure[Existing IL methods.]{
\begin{minipage}[t]{0.46\linewidth}
\centering
\includegraphics[width=\linewidth]{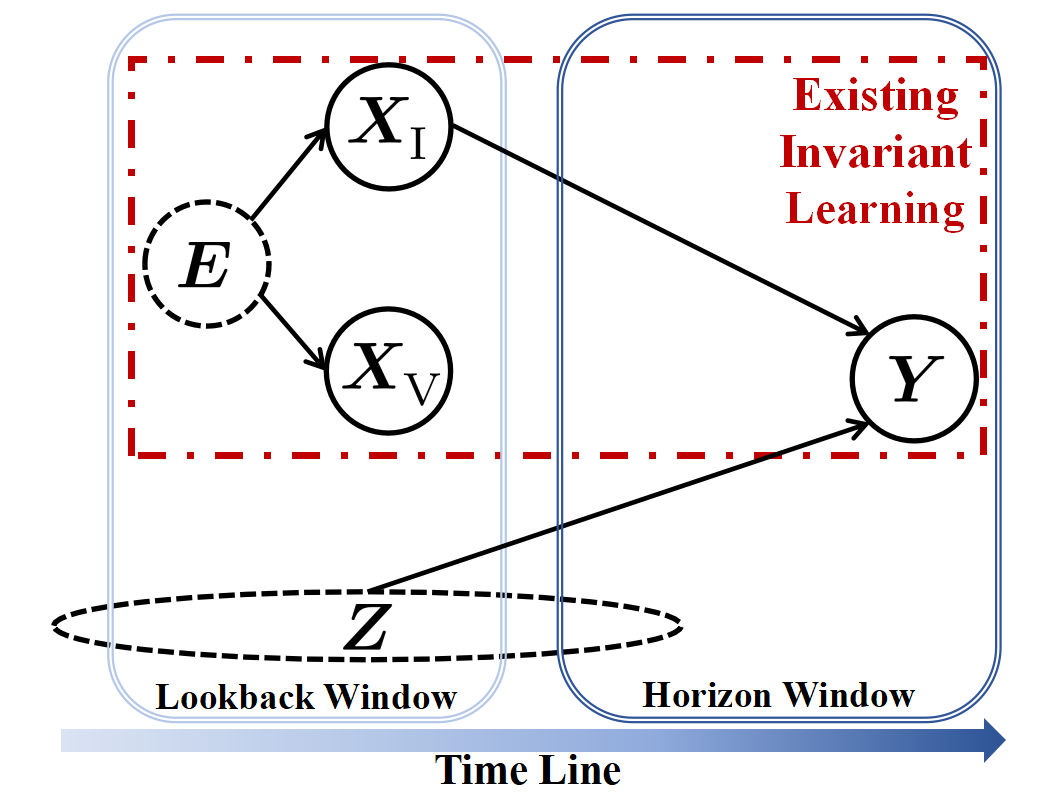}
\label{fig:CG_A}
\end{minipage}
}\hspace{1mm}
\subfigure[Our proposed method.]{
\begin{minipage}[t]{0.46\linewidth}
\centering
\includegraphics[width=\linewidth]{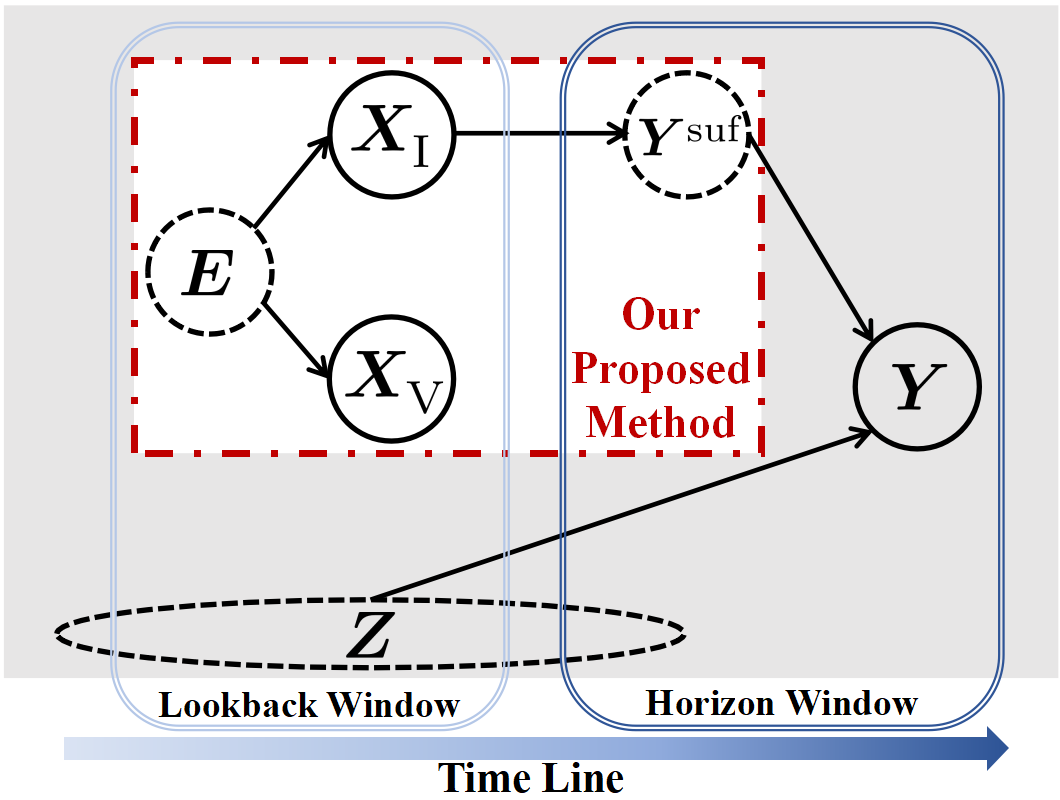}
\label{fig:CG_B}
\end{minipage}
}
\caption{ The structural causal model (SCM) for (a) existing invariant learning methods and (b) our proposed method.
The key difference is that our method targets the sufficiently predictable part of the target, i.e., \(\bm{\YSUF}\) rather than the raw \(\bm{Y}\), thus making invariant learning feasible.}
\end{figure}

In the SCM shown in Figure~\ref{fig:CG_A}, we use $\bm{Z} \rightarrow \bm{Y}$ and the dash circle to describe the core effect of $\bm{Z}$ on $\bm{Y}$ and the unobserved issue of $\bm{Z}$ respectively. 
Clearly, there exists a gap between the SCM modeled by the existing IL methods and the SCM underlying TS data , due to the existence of $\bm{Z}$.

The existence of unobserved $\bm{Z}$ breaks both two parts of the IL's conventional assumption~\ref{assump:fund}: First, $\bm{Z}$ breaks the sufficiency property part, obviously. 
Thus, existing IL methods actually absorb the influence of $\bm{Z}$ on $\bm{Y}$, leading to the overfitting issue, especially with deep models.
Second, $\bm{Z}$ breaks the invariance property part when $\bm{Z}$ and $\bm{E}$ are not independent, for example, influenza outbreaks occur more frequently in winter.
Formally, if there exists $ e_i, e_j \in \supportE$ such that $P^{e_{i}}(\bm{Z}|\XI) \neq P^{e_{j}}(\bm{Z}|\XI)$, then we have $P^{e_{i}}(\bm{Y}|\XI) = \sum_{\mathbf{Z}} P(\bm{Y}|\XI, \mathbf{Z})P^{e_{i}}(\mathbf{Z}|\XI) \neq P^{e_{j}}(\bm{Y}|\XI)$.
Thus, existing IL methods lacks reliable OOD generalization ability for TSF.

\par \noindent \textbf{TS datasets usually lack environment labels.}
Firstly, most IL methods~\cite{IRM,IB-ERM,VREx,SD,GroupDRO} require explicit environment labels as input, which are often unavailable in TSF datasets.
Due to the complexity of temporal environments, manual annotation is often difficult, expensive, and sometimes suboptimal.
Secondly, existing IL with environment inference methods are fundamentally not applicable for TSF: (1) Existing IL methods show certain limitations when applying to TSF tasks:
HRM~\cite{HRM} and KernelHRM~\cite{khrm} are based on low-dimensional raw features, while TS data are typically high-dimensional; EIIL~\cite{EIIL} needs delicate initialization; ZIN~\cite{ZIN} requires additional information satisfying specific conditions; and EDNIL~\cite{mhead} is designed for classification tasks. 
(2) Existing IL methods primarily cater to static data and thus overlook the characteristics of time-series data, leading to suboptimal inferred environments.

\section{Our Methodology}\label{sec:method}
We propose \textbf{\ourmethod} (\underline{F}orcasting for-\underline{O}ut-of-distribution generalization via~\underline{I}nvariant~\underline{L}earning), a model-agnostic environment-aware invariant learning framework, serving as a practical solution for the OOD-TSF problem. 
\subsection{Overview}
\par \noindent \textbf{High-level Idea.}
Our main idea is to use IL with environment inference targeting at the sufficiently predictable part of the target (we call it $\YSUF$), see Figure~\ref{fig:CG_B}.
Specifically, inspired by the Wold's decomposition theorem~\cite{w1,w2}, we assume that $\bm{Y}$ can be decomposed into deterministic and uncertain parts relative to the input $\bm{X}$. Formally, $\bm{Y} = q(\YSUF, \bm{Z})$, with $q(\cdot,\cdot)$ as any mapping function. Here, $\YSUF\in\mathcal{Y}$, determined by the input $\bm{X}$, is deterministic, i.e., sufficiently predictable. Thus, targeting at $\YSUF$, the Assumption~\ref{assump:fund} of sufficiency and invariance property holds, making invariant learning feasible. Additionally, considering the unpredictability of unobserved $\bm{Z}$, the optimal OOD prediction can be achieved if we are able to uncover $\YSUF$ via invariant features $\XI$. To this end, we propose FOIL, which serves as a practical solution for applying IL to the OOD-TSF problem.
\par
\begin{figure}[t]
\centering
\includegraphics[width=\linewidth]{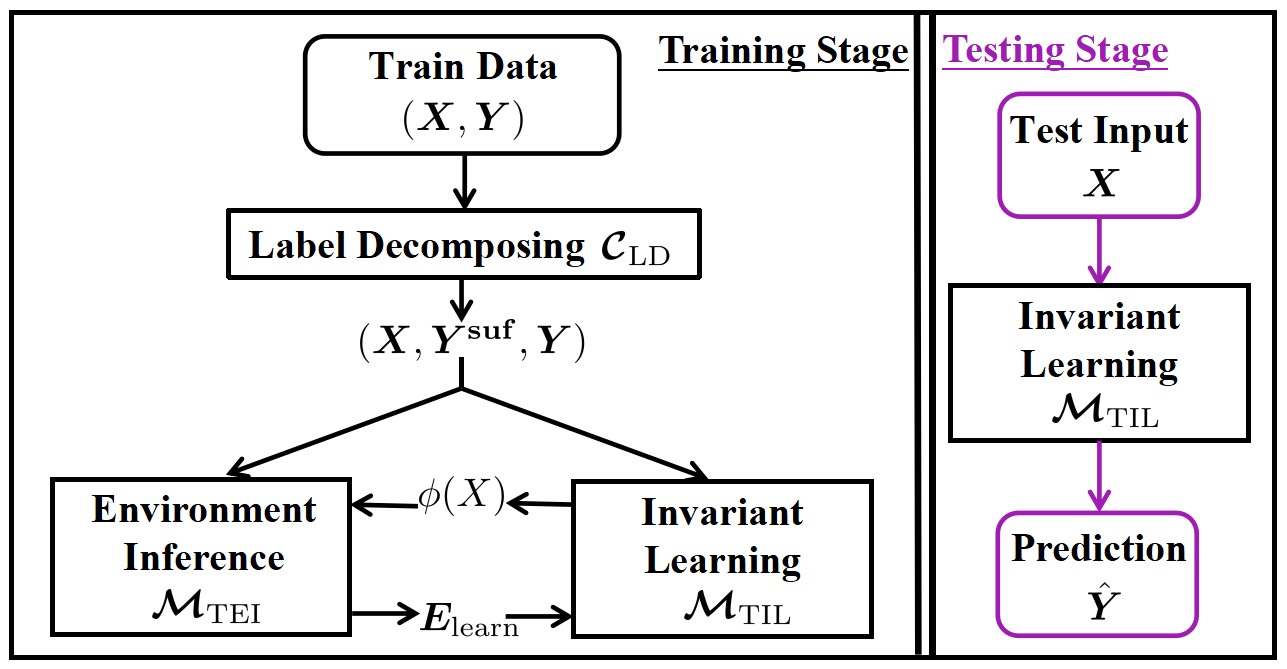}
\caption{The overall framework of our proposed \ourmethod.}
\label{fig:model_c}
\end{figure}

 \noindent \textbf{Overall Framework.}
As shown in Figure~\ref{fig:model_c}, \ourmethod consists of three parts:

(1) \textit{Label Decomposing Component} ($\CLD$), which decomposes sufficiently predictable $\YSUF$ from observed $\bm{Y}$.

(2) \textit{Time-Series Environment Inference Module} ($\MTEI$), which infers temporal environments based on learned representations from $\MTIL$.

(3) \textit{Time-Series Invariant Learning Module} ($\MTIL$), which learns invariant representations across inferred environments from $\MTEI$.

In \ourmethod, $\CLD$ is the preliminary step for $\MTIL$ and $\MTEI$; $\MTIL$ and $\MTEI$ are then jointly optimized via alternating updates. During the testing phase, only $\MTIL$ is utilized for prediction. 

As the first work of IL for TSF, \ourmethod is designed as a model-agnostic framework that seamlessly incorporates various off-the-shelf deep TSF models. 
Specifically, the backbone model can be \textit{any deep TSF model}, denoted $\phi(\boldsymbol{X})$.
We append a regressor $\rho(\cdot)$, typically a fully connected layer, on top of the learned output representations
from the backbone model $\phi(\cdot)$.
We denote the combined model succinctly as
as $f_\theta(\bm{X})=\rho\left(\phi(\boldsymbol{X})\right)$.
$\MTIL$ and $\MTEI$ leverage the output representation $\phi(\boldsymbol{X})$, both for achieving model-agnostic and for accommodating high-dimensional inputs of TSF. We will next introduce each part.

\subsection{The Label Decomposing Component}\label{section:CLD}
$\CLD$ is used to decompose the sufficiently predictable $\YSUF$ from the observed $\bm{Y}$. 
However, accurately obtaining \(\YSUF\) is nearly unfeasible, owing to the lack of information about the underlying generation function and unobserved variables $\bm{Z}$. 
Instead of introducing additional data, such as external datasets as the agent for \(\bm{Z}\), we aim to alleviate this problem more practically via a surrogate loss to mitigate the effect of $\bm{Z}$. 
Firstly, we add the following assumption: 
\begin{equation}
    \bm{Y} = q( \YSUF,  \bm{Z})=\alpha(\bm{Z})(\YSUF) + \beta(\bm{Z})\mathbf{1},
\label{assump2}
\end{equation}
where $\alpha(\cdot): \mathbb{R}^{d_Z} \to \mathbb{R}$ and $\beta(\cdot): \mathbb{R}^{d_Z} \to \mathbb{R}$ could be any mapping function, and $\mathbf{1} \in \mathbb{R}^{h\times d_{\mathrm{out}}}$ is an all-one matrix. This assumption follows the dynamic nature of observed $\bm{Y}$'s distribution~\cite{cite1}.
Specifically, this assumption encompasses two aspects:

(1) The relationships between $\bm{Z}$ and $\YSUF$ are additive and multiplicative, which is a widely adopted assumption about unobserved variables~\cite{add1,add2,add01,add02}. (2) $\bm{Z}$ exerts a consistent influence in one horizon window, which can be readily extended by partitioning the horizon window into multiple segments. 
Thus, the residual $\bm{Res}$ between ground truth $\bm{Y}$ and predicted $\hat{\bm{Y}}$, i.e., $\bm{Res} = \bm{Y} - \hat{\bm{Y}}$, absorb the effect of $\bm{Z}$ on $\bm{Y}$ via values of mean $\mu(\bm{Res})$ and standard deviation $\sigma(\bm{Res})$. Thus, we propose an Instance Residual Normalization (IRN) method to mitigate the effect of $\bm{Z}$. For the residual $\bm{Res}_t$ of instance $t$, IRN method can be formulated as:
\begin{equation}
    \tilde{\mathbf{Res}}_{t} = \frac{\mathbf{Y}_t -  \mu\left(\mathbf{Y}_t \right)}{\sigma(\mathbf{Y}_t)} - \frac{\hat{\mathbf{Y}}_t - \mu\left( \hat{\mathbf{Y}}_t\right)}{\sigma(\hat{\mathbf{Y}}_t)}= \tilde{\mathbf{Y}}_t - \tilde{\hat{\mathbf{Y}}}_t
\label{eq:residualnorm}
\end{equation}
IRN in Eq.~\ref{eq:residualnorm} ensures the residuals to have a mean of 0 and a variance of \(2 - 2\mathrm{cov}(\hat{\mathbf{Y}}, \mathbf{Y})\), where \(\mathrm{cov}\) denotes the covariance. 

Finally, we derive the following simple and effective surrogate loss to mitigate the effect of $\bm{Z}$, instead of directly decoupling $\YSUF$ in $\CLD$:
\begin{equation}
    \LSUF(\hat{\bm{Y}}, \bm{Y})= \mathrm{MSE}(\tilde{\bm{Res}}, \bm{0})= \ell(\tilde{\hat{\bm{Y}}}, \tilde{\bm{Y}}),
\label{eq:loss_suf}
\end{equation}
where $\mathrm{MSE}(\tilde{\bm{Res}}, \bm{0})=\frac{1}{h} \sum_{j=1}^{h} (\tilde{\mathbf{Res}}_{t+j})^2$. Note that our IRN fundamentally differs from the existing instance normalization (IN) methods. Existing methods adopt IN to \(\bm{X}\), and reverse IN to \(\hat{\bm{Y}}\) based on $\mu(\bm{X})$ and $\sigma(\bm{X})$, aiming to address non-stationary problem of $\bm{X}$~\cite{revin,Nonstationary}. 
While, our IRN method directly aligns the mean and variance between \(\hat{\bm{Y}}=f(\bm{X})\) and \(\bm{Y}\), thus removing error caused by \(\bm{Z}\) under the introduced assumption. Since \(\bm{Z}\) is not contained in \(\bm{X}\), existing methods usually fail to achieve our goal.

\subsection{The Time-Series Environment Inference Module}\label{Sec:TEI}
$\MTEI$ aims to infer environments $\bm{E}_\mathrm{infer}$, thereby providing environment labels for the time-series invariant learning module $\MTIL$. 
We consider inferring effective and reasonable temporal environments with two goals:
\par \noindent \textbf{(1) Sensitive to the encoded invariant features.} In \ourmethod, $\MTEI$ and $\MTIL$ are adversarial: $\MTEI$ infers environments based on the variant features not discarded by $\MTIL$; $\MTIL$ discards variant features based on inferred environments from $\MTEI$. Ultimately, when $\MTIL$ only utilizes invariant features, $\MTEI$ is unable to infer effective environments. 
Thus, we propose to infer informative environments that are sensitive to the variant features encoded in the currently learned representations, formulated as:
\begin{equation}
    \min_{\bm{E}_\mathrm{infer}} H\left(\YSUF|\phi^*(\bm{X}),\bm{E}_\mathrm{infer}\right)- H\left(\YSUF|\phi^*(\bm{X})\right),
\label{EI:formulation}
\end{equation}
where $H$ is Shannon conditional entropy, $\phi^{*}(\bm{X})$ are learned representations from $\MTIL$ and frozen in $\MTEI$.
\par \noindent \textbf{(2) Preserving the temporal adjacency structures.} To ensure that the inferred environments are reasonable in the context of TSF, we consider preserving the inherent characteristic of time-series data, i.e., the temporal adjacency structure. 
Specifically, instances that are temporally adjacent should possess similar temporal environments. This can also be viewed as a type of regularization to prevent inferred environments from overfitting to random noises.

Intuitively, the approach to infer environments is to optimize Eq.~\ref{EI:formulation}, with a plugin for preserving the temporal adjacency structure. To this end, we present an EM-based clustering solution in the representation space, implemented through a multi-head neural network. Each head is an environment-specific regressor, playing the role of each cluster's center. Specifically, the regressor $\rho^{(e)}$ is specific for environment $e$. And the representation $\phi^{*}(\bm{X})$ is shared and frozen in $\MTEI$. We describe the M step and E step next.
\paragraph{M Step: Optimizing Environment-Specific Regressors}
\mbox{}\\
In the M step, we optimize $\{\rho^{(e)}\}$ to better fit the data from current environment partition $\bm{E}_{\mathrm{infer}}$ of E step as:
\begin{equation}
\begin{aligned}
    &\min_{\{\rho^{(e)}\}} 
\bm{\mathcal{L}}_{\mathrm{TEI}}
=\mathbb{E}_{e \in \bm{E}_{\mathrm{infer}}}\bm{\mathcal{R}}^{(e)}_{\mathrm{suf}}(\rho^{(e)},\phi^{*}) \\
&=\sum_{e \in \bm{E}_{\mathrm{infer}}} \frac{1}{|\mathcal{D}_e|} \sum_{(\mathbf{X},\mathbf{Y}) \in \mathcal{D}_e} \LSUF\left(\rho^{(e)}\left(\phi^*(\mathbf{X})\right),\mathbf{Y}\right)
\end{aligned}
\label{loss:TEI}
\end{equation}
\paragraph{E Step: Estimating Environment Labels}
\mbox{}\\
 Next, in the E step, we reallocate the environment partitions. For instance $(\mathbf{X}_{t},\mathbf{Y}_t)$, we reassign its environment label $\bm{E}_{\mathrm{infer}}(t)$ via the following two steps: 
\begin{itemize}
\setlength{\itemsep}{0pt}
    \item Step 1: Reallocating based on the distances with the center of each cluster (environment). We use the loss with respect to regressor $\rho^{(e)}$ to describe the distance with the center of cluster $e$. Thus, we reassign $\bm{E}_{\mathrm{infer}}(t)$ according to the shortest distance, as follows:
    \begin{equation}
    \bm{E}_{\mathrm{infer}}(t) \leftarrow \arg \underset{e \in \bm{E}_{\mathrm{infer}}}{\min} \left\{\LSUF\left(\rho^{(e)}\left(\phi^*(\mathbf{X}_{t})\right),\mathbf{Y}_{t}\right)\right\}
    \label{eq:EP}
    \end{equation}
    \item Step 2: Reallocating to preserve temporal adjacency structure. We propose an environment label propagation solution to achieve this goal, as follows:
    \begin{equation}
    \bm{E}_{\mathrm{infer}}(t) \leftarrow \mathrm{mode}\left\{\bm{E}_{\mathrm{infer}}(t+j)\right\}_{j=-r}^{r},
    \label{eq:LP} 
    \end{equation}
    where mode implements majority voting by considered temporal neighbors selected via the radius $r \in \mathbb{Z}^+$.
\end{itemize}
In summary, we iteratively execute M step and E step to obtain the inferred environments \(\bm{E}^{*}_{\mathrm{infer}}\). Due to the fixed second term of Eq.~\ref{EI:formulation}, our solution represents a practical instantiation of Eq.~\ref{EI:formulation}.

\subsection{The Time-Series Invariant Learning Module}
$\MTIL$ is used to learn invariant representations $\phi^{*}(\bm{X})$ across inferred environments $\bm{E}^{*}_{\mathrm{{infer}}}$ from $\MTEI$. Specifically, $\MTIL$ aims to learn the $\phi^*({\bm{X}})$ which encode and solely encode all the information of invariant features $\XI$ thus achieving both invariant and sufficient predictive capability targeting at \(\YSUF\). Such $\phi^*(X)$ has been theoretically proven \cite{HRM} to be obtained by optimizing the following objective function:
\begin{equation}
    \phi^*=\arg \max_{\phi}I(\mathbf{Y}^{\mathrm{suf}};\phi(\mathbf{X}) - I(\mathbf{Y}^{\mathrm{suf}};\bm{E}^{*}_{\mathrm{learn}}|\phi(\mathbf{X})),
\label{IL:obj}
\end{equation}
where $I(\cdot;\cdot)$ measures Shannon mutual information. The first and second terms correspond to ensure sufficiency and invariance property of $\phi(\bm{X})$, respectively. 

Considering the unavailability of $\YSUF$, we present the following practical loss function as the instantiation of Eq.~\ref{IL:obj} via our surrogate loss in Eq.~\ref{eq:loss_suf}:
\begin{equation}
\begin{aligned}
       \min_{\rho,\phi}\mathcal{L}_{\mathrm{TIL}}= &\mathbb{E}_{e \in \bm{E}^{*}_{\mathrm{infer}}}\bm{\mathcal{R}}^{(e)}_{\mathrm{suf}}(\rho,\phi)+\lambda_1\bm{\mathcal{R}}_{\mathrm{ERM}}(\rho,\phi)\\
        &+\lambda_2\mathrm{Var}_{e\in{\bm{E}^{*}}_{\mathrm{infer}}}\left[\bm{\mathcal{R}}^{(e)}_{\mathrm{suf}}(\rho,\phi)\right], 
\end{aligned}
\label{loss:IL}
\end{equation}
where $\lambda_1, \lambda_2$ are hyper-parameters, $\bm{\mathcal{R}}_{\mathrm{ERM}}(\rho, \phi) = \mathbb{E}_{\mathbf{X}, \mathbf{Y}}\left[\ell(\rho(\phi(\mathbf{X})), \mathbf{Y})\right]$ is the ERM loss on raw $\bm{Y}$, $\bm{\mathcal{R}}^{e}_{\mathrm{suf}}(\rho, \phi)$ defined in Eq.~\ref{loss:IL} is the  loss of inferred environment $e$ on $\YSUF$, and $\mathrm{Var}_{e \in \bm{E}^{*}_{\mathrm{infer}}}\left[\bm{\mathcal{R}}^{(e)}_{\mathrm{suf}}(\rho, \phi)\right]$ implies the variance of loss across inferred environments. The first and second terms are jointly used to ensure the sufficient predictive power of $\phi(\bm{X})$ for $\YSUF$, where $\lambda_1$ controls the trade-off between introducing information of $\mu(\YSUF), \sigma(\YSUF)$ and the influence of $\bm{Z}$. The third term further balanced by $\lambda_2$ ensures the invariance property and is robust to marginal distribution shifts of input, theoretically guaranteed by \cite{VREx} and further balanced by $\lambda_2$.

The overall algorithm is summarized in Appendix~\ref{AP:Al}. Compared to the backbone, \ourmethod slightly increases the parameter count due to additional multiple regressors. 
\begin{table*}[ht]
\centering
\renewcommand{\arraystretch}{1}
\resizebox{\textwidth}{!}{%
\begin{tabular}{cccccccccccccc}
\hline
\multicolumn{2}{c}{Method} & \multicolumn{2}{c}{Informer(AAAI'21)} & \multicolumn{2}{c}{with \ourmethod} & \multicolumn{2}{c}{Crossformer(ICLR'23)} & \multicolumn{2}{c}{with \ourmethod} & \multicolumn{2}{c}{PatchTST(ICLR'23)} & \multicolumn{2}{c}{with \ourmethod} \\ \hline
\multicolumn{2}{c|}{Metric} & MSE & MAE & MSE & MAE & MSE & MAE & MSE & MAE & MSE & MAE & MSE & MAE \\ \hline
\multicolumn{1}{c|}{\multirow{7}{*}{\rotatebox[origin=c]{270}{Exchange}}} & 
\multicolumn{1}{c|}{24} & 0.812 & 0.736 & {\textbf{0.036}} & \multicolumn{1}{c|}{\textbf{0.146}} & 0.083 & 0.233 & \textbf{0.029} & \multicolumn{1}{c|}{\textbf{0.129}} & 0.092 & 0.229 & {\textbf{0.031}} & {\textbf{0.136}} \\
\multicolumn{1}{c|}{} & \multicolumn{1}{c|}{48}  & 0.715 & 0.682 & {\textbf{0.063}} & \multicolumn{1}{c|}{\textbf{0.191}} & 0.164 & 0.328 & \textbf{0.054} & \multicolumn{1}{c|}{\textbf{0.175}} & 0.090 & 0.243 & {\textbf{0.052}} & {\textbf{0.171}} \\
\multicolumn{1}{c|}{} & \multicolumn{1}{c|}{96}  & 0.782 & 0.710 & {\textbf{0.142}} & \multicolumn{1}{c|}{\textbf{0.274}} & 0.214 & 0.381 & \textbf{0.111} & \multicolumn{1}{c|}{\textbf{0.240}} & 0.142 & 0.291 & {\textbf{0.107}} & {\textbf{0.235}}
 \\
\multicolumn{1}{c|}{} & \multicolumn{1}{c|}{192} & 0.708 & 0.701 & {\textbf{0.236}} & \multicolumn{1}{c|}{\textbf{0.369}} & 0.709 & 0.716 & \textbf{0.213} & \multicolumn{1}{c|}{\textbf{0.349}} & 0.364 & 0.468 & {\textbf{0.226}} & {\textbf{0.351}} \\
\multicolumn{1}{c|}{} & \multicolumn{1}{c|}{336}  & 1.587 & 1.063 & {\textbf{0.546}} & \multicolumn{1}{c|}{\textbf{0.591}} & 2.158 & 1.231 & \textbf{0.471} & \multicolumn{1}{c|}{\textbf{0.500}} & 0.512 & 0.540 & {\textbf{0.465}} & {\textbf{0.486}}\\
\multicolumn{1}{c|}{} & \multicolumn{1}{c|}{720}   & 3.922 & 1.793 & {\textbf{0.712}} & \multicolumn{1}{c|}{\textbf{0.679}} & 2.093 & 1.215 & \textbf{1.193} & \multicolumn{1}{c|}{\textbf{0.833}} & 0.957 & 0.738 & {\textbf{0.925}} & {\textbf{0.722}}\\
\multicolumn{1}{c|}{} & \multicolumn{1}{c|}{IMP.} &               &              & \textbf{80.58\%}     & \multicolumn{1}{c|}{\textbf{61.24\%}}    &                &                & \textbf{61.90\%}    & \multicolumn{1}{c|}{\textbf{45.06\%}}    &               &              & \textbf{30.60\%}             & \textbf{21.11\%}            \\ 
\hline
\multicolumn{1}{c|}{\multirow{7}{*}{\rotatebox[origin=c]{270}{ILI}}} 
& \multicolumn{1}{c|}{4}& 3.212 & 1.530 & {\textbf{0.736}} & \multicolumn{1}{c|}{\textbf{0.593}} & 2.147 & 1.232 & \textbf{0.332} & \multicolumn{1}{c|}{\textbf{0.400}} & 1.043 & 0.587 & {\textbf{0.616}} & {\textbf{0.507}} \\
\multicolumn{1}{c|}{} & \multicolumn{1}{c|}{8} & 3.668 & 1.642 & {\textbf{0.881}} & \multicolumn{1}{c|}{\textbf{0.667}} & 2.678 & 1.403 & \textbf{0.569} & \multicolumn{1}{c|}{\textbf{0.512}} & 0.638 & 0.557 & {\textbf{0.586}} & {\textbf{0.546}} \\
\multicolumn{1}{c|}{} & \multicolumn{1}{c|}{12}     & 3.974 & 1.722 & {\textbf{1.069}} & \multicolumn{1}{c|}{\textbf{0.768}} & 2.914 & 1.476 & \textbf{0.706} & \multicolumn{1}{c|}{\textbf{0.575}} & 0.959 & 0.795 & {\textbf{0.560}} & {\textbf{0.519}} \\
\multicolumn{1}{c|}{} & \multicolumn{1}{c|}{16} & 4.187 & 1.773 & {\textbf{1.047}} & \multicolumn{1}{c|}{\textbf{0.779}} & 3.496 & 1.628 & \textbf{0.701} & \multicolumn{1}{c|}{\textbf{0.568}} & 0.726 & 0.563 & {\textbf{0.696}} & {\textbf{0.555}} \\
\multicolumn{1}{c|}{} & \multicolumn{1}{c|}{20} & 4.296 & 1.806 & {\textbf{1.011}} & \multicolumn{1}{c|}{\textbf{0.797}} & 3.589 & 1.653 & \textbf{0.702} & \multicolumn{1}{c|}{\textbf{0.596}} & 0.807 & 0.705 & {\textbf{0.571}} & {\textbf{0.541}}\\
\multicolumn{1}{c|}{} & \multicolumn{1}{c|}{24} & 4.445 & 1.844 & {\textbf{1.014}} & \multicolumn{1}{c|}{\textbf{0.806}} & 3.513 & 1.633 & \textbf{0.686} & \multicolumn{1}{c|}{\textbf{0.604}} & 1.072 & 0.850 & {\textbf{0.663}} & {\textbf{0.625}} \\
\multicolumn{1}{c|}{} & \multicolumn{1}{c|}{IMP.} &  &  &\textbf{75.80\%}     & \multicolumn{1}{c|}{\textbf{57.37\%}}    &                &                & \textbf{79.99\%}    & \multicolumn{1}{c|}{\textbf{64.03\%}}    &               &              &\textbf{27.04\%}             &\textbf{16.91\%}          \\
\hline
\multicolumn{1}{c|}{\multirow{7}{*}{\rotatebox[origin=c]{270}{ETTh1}}} & 
\multicolumn{1}{c|}{24} & 0.219 & 0.392 & {\textbf{0.038}} & \multicolumn{1}{c|}{\textbf{0.146}} & 0.194 & 0.400 & \textbf{0.028} & \multicolumn{1}{c|}{\textbf{0.126}} & 0.031 & 0.136 & {\textbf{0.027}} & {\textbf{0.126}} \\
\multicolumn{1}{c|}{} & \multicolumn{1}{c|}{48} & 0.474 & 0.638 & {\textbf{0.065}} & \multicolumn{1}{c|}{\textbf{0.190}} & 0.270 & 0.465 & \textbf{0.042} & \multicolumn{1}{c|}{\textbf{0.156}} & 0.044 & 0.160 & {\textbf{0.041}} & {\textbf{0.154}} \\
\multicolumn{1}{c|}{} & \multicolumn{1}{c|}{96} & 0.965 & 0.892 & {\textbf{0.088}} & \multicolumn{1}{c|}{\textbf{0.224}} & 0.146 & 0.312 & \textbf{0.056} & \multicolumn{1}{c|}{\textbf{0.181}} & 0.061 & 0.190 & {\textbf{0.056}} & {\textbf{0.182}} \\
\multicolumn{1}{c|}{} & \multicolumn{1}{c|}{192} & 1.029 & 0.967 & {\textbf{0.148}} & \multicolumn{1}{c|}{\textbf{0.299}} & 0.241 & 0.420 & \textbf{0.075} & \multicolumn{1}{c|}{\textbf{0.209}} & 0.082 & 0.223 & {\textbf{0.078}} & {\textbf{0.215}} \\
\multicolumn{1}{c|}{} & \multicolumn{1}{c|}{336}     & 0.677 & 0.769 & {\textbf{0.136}} & \multicolumn{1}{c|}{\textbf{0.296}} & 0.246 & 0.425 & \textbf{0.088} & \multicolumn{1}{c|}{\textbf{0.233}} & 0.100 & 0.246 & {\textbf{0.092}} & {\textbf{0.237}}
 \\
\multicolumn{1}{c|}{} & \multicolumn{1}{c|}{720}& 1.086 & 0.973 & {\textbf{0.132}} & \multicolumn{1}{c|}{\textbf{0.288}} & 0.392 & 0.554 & \textbf{0.104} & \multicolumn{1}{c|}{\textbf{0.254}} & 0.154 & 0.310 & {\textbf{0.120}} & {\textbf{0.272}} \\
\multicolumn{1}{c|}{} & \multicolumn{1}{c|}{IMP.} &  &  &\textbf{85.38\%}     & \multicolumn{1}{c|}{\textbf{68.14\%}}    &                &                & \textbf{73.04\%}    & \multicolumn{1}{c|}{\textbf{54.43\%}}    &               &              &\textbf{10.48\%}             &\textbf{5.80\%}            \\ 
\hline
\multicolumn{1}{c|}{\multirow{7}{*}{\rotatebox[origin=c]{270}{ETTh2}}} & 
\multicolumn{1}{c|}{24}     & 0.668 & 0.705 & {\textbf{0.121}} & \multicolumn{1}{c|}{\textbf{0.275}} & 0.136 & 0.299 & \textbf{0.071} & \multicolumn{1}{c|}{\textbf{0.198}} & 0.080 & 0.215 & {\textbf{0.071}} & {\textbf{0.197}}
 \\
\multicolumn{1}{c|}{} & \multicolumn{1}{c|}{48}    & 0.999 & 0.866 & {\textbf{0.258}} & \multicolumn{1}{c|}{\textbf{0.407}} & 0.122 & 0.274 & \textbf{0.106} & \multicolumn{1}{c|}{\textbf{0.248}} & 0.106 & 0.248 & {\textbf{0.103}} & {\textbf{0.241}}
 \\
\multicolumn{1}{c|}{} & \multicolumn{1}{c|}{96}     & 3.070 & 1.628 & {\textbf{0.222}} & \multicolumn{1}{c|}{\textbf{0.369}} & 0.256 & 0.408 & \textbf{0.137} & \multicolumn{1}{c|}{\textbf{0.286}} & 0.156 & 0.309 & {\textbf{0.140}} & {\textbf{0.289}}
 \\
\multicolumn{1}{c|}{} & \multicolumn{1}{c|}{192}     & 3.548 & 1.768 & {\textbf{0.699}} & \multicolumn{1}{c|}{\textbf{0.682}} & 1.257 & 1.034 & \textbf{0.198} & \multicolumn{1}{c|}{\textbf{0.352}} & 0.217 & 0.374 & {\textbf{0.201}} & {\textbf{0.356}}
 \\
\multicolumn{1}{c|}{} & \multicolumn{1}{c|}{336}     & 2.663 & 1.526 & {\textbf{0.801}} & \multicolumn{1}{c|}{\textbf{0.756}} & 1.305 & 1.027 & \textbf{0.234} & \multicolumn{1}{c|}{\textbf{0.389}} & 0.233 & 0.390 & {\textbf{0.216}} & {\textbf{0.372}}
 \\
\multicolumn{1}{c|}{} & \multicolumn{1}{c|}{720}     & 2.335 & 1.422 & {\textbf{0.730}} & \multicolumn{1}{c|}{\textbf{0.725}} & 1.579 & 1.158 & \textbf{0.253} & \multicolumn{1}{c|}{\textbf{0.402}} & 0.317 & 0.448 & {\textbf{0.238}} & {\textbf{0.391}}
 \\
\multicolumn{1}{c|}{} & \multicolumn{1}{c|}{IMP.} &  &  &\textbf{78.03\%}     & \multicolumn{1}{c|}{\textbf{58.82\%}}    &                &                & \textbf{59.61\%}    & \multicolumn{1}{c|}{\textbf{44.42\%}}    &               &              &\textbf{10.65\%}              &\textbf{6.64\%}            \\ 
\hline
\end{tabular}%
}
\caption{Performance comparison between original and \ourmethod equipped versions of backbones. The top-performing version is marked in \textbf{bold}. IMP. is the average percentage improvement across lengths of horizon window compared to the original version. \ourmethod consistently and significantly enhances the performance of various TSF backbones on all datasets and metrics across horizon window lengths.}
\label{tab:overall_camp}
\end{table*}
\section{Experiments}\label{sec:exp}
\subsection{Setup}
 \par \noindent \textbf{Datasets.} We conduct experiments on four popular real-world datasets commonly used as benchmarks: the daily reported exchange rates dataset (\textbf{Exchange})~\cite{exc}, the weekly reported ratios of patients seen with influenza-like illness dataset (\textbf{ILI})~\cite{I2}, and two hourly reported electricity transformer temperature datasets (\textbf{ETTh1} and \textbf{ETTh2})~\cite{informer}. We adhere to the general setups and target variables selections, following previous literatures~\cite{autoformer,timesnet,patchtst}.

\par \noindent \textbf{Backbones.} As previously mentioned, our proposed \ourmethod is a model-agnostic framework. We select three different types of TSF models as backbones. \textbf{Informer}~\cite{informer} proposes an efficient transformer for long-term TSF. \textbf{Crossformer}~\cite{crossformer} better utilizes cross-dimension dependency, making it more sensitive to spuriouse correlations. \textbf{PatchTST}~\cite{patchtst} employs channel-independent and patching strategies to achieve state-of-the-art performance.

\par \noindent \textbf{Baselines.} We comprehensively compare the following twelve distribution shifts baselines: (1) Two advanced methods for handling temporal distribution shifts in TSF: \textbf{NST}~\cite{Nonstationary} and \textbf{RevIN}~\cite{revin}. (2) Six well-acknowledged general OOD methods following~\cite{woods}, adopted due to the lack of OOD methods for TSF: (a)  Methods requiring environment labels: \textbf{GroupDRO}~\cite{GroupDRO}, \textbf{IRM}~\cite{IRM}, \textbf{IB-ERM}~\cite{IB-ERM}, \textbf{VREx}~\cite{VREx} and \textbf{SD}~\cite{SD}. (b)  Methods not requiring environment labels: \textbf{EIIL}~\cite{EIIL}. (3) Two hybrid methods: \textbf{IRM+RevIN} and \textbf{EIIL+RevIN}.

\par \noindent \textbf{Implementation.} Regarding the horizon window length, we considered a range from short to long-term TSF tasks. For ETTh1, ETTh2, and Exchange, the lengths are [24, 48, 96, 168, 336, 720] with a fixed lookback
window size of 96 and a consistent label window size of 48 for the decoder. For the weekly reported ILI, the lengths are [4, 8, 12, 16, 20, 24], representing the next one month to six months, with a fixed lookback
window size of 36 and a consistent label window size of 18 for the decoder.. Note that, we lack the availability of suitable environment labels. We address this by dividing the training set into \( k \), tuned from 2 to 10, equal-length time segments to serve as predefined environment labels. When the backbone is equipped with our \ourmethod, the model architecture of the backbone remains unchanged.

\par \noindent \textbf{Evaluation.} We employ the widely-adopted evaluation metrics: mean squared error (\textbf{MSE}) and mean absolute error (\textbf{MAE}). We report average performance over three independent runs for each model.

\par \noindent \textbf{Reproducibility.} All training data, testing data and code are available at: \url{https://github.com/AdityaLab/FOIL}. More experimental details are revealed in Appendix~\ref{exp}.

\subsection{Results}
As shown in Table~\ref{tab:overall_camp}, we present results for both original versions and corresponding \ourmethod equipped versions of backbones, yielding the following observations:
\par \noindent (1) Overall, \ourmethod consistently and significantly improves the performance of various TSF backbones across all datasets and forecasting lengths with improvements reaching up to 85\% on MSE, thereby demonstrating \ourmethod's effectiveness. For the state-of-the-art PatchTST, \ourmethod consistently enhances performance, achieving up to 30\% improvement.
For the lower-performing Informer, \ourmethod shows more significant improvements, frequently by an order of magnitude, yielding competitive results.
\par \noindent (2) \ourmethod excels in short-term forecasting compared to long-term forecasting, as the higher uncertainty of the latter hinders learning invariant features. 
Moreover, \ourmethod's most significant improvement in the ILI dataset is attributed to the serious OOD shifts in its test data, particularly during the unseen COVID-19 period.

\subsection{Comparison with Distribution Shifts Methods}\label{exp:cmp}
In this section, we conduct a comparative analysis of the performance disparities between \ourmethod and existing distribution shifts methods. We employ the Informer as the forecasting backbone.
The forecasting length is set as 16 for ILI and 96 for others.
Similar observations are found in other settings. We measure the relative improvement compared to the best-performing baseline on each metric and dataset.

As shown in Table~\ref{tab:OOD_compar}, our observations include:

(1) \ourmethod achieves the best performance across all datasets. The average improvements on MSE and MAE are more than 10\% and 5.5\% respectively, showing the benefits of \ourmethod over existing distribution shift methods. Notably, though hybrid models additionally alleviate the temporal distribution shift problem and exhibit better performance than general OOD baselines, \ourmethod still outperforms hybrid models by over 11\%. Therefore, our proposed surrogate loss in Eq.~\ref{eq:loss_suf} is irreplaceable by current instance normalization methods as discussed in Section~\ref{section:CLD} and exhibits important benefits for alleviating unobserved core covariates issues in the TSF task. 

(2) General OOD methods exhibit poor performances. This verifies that directly applying existing invariant learning methods for the TSF task is inappropriate, as discussed in Section~\ref{sec:gap}. 

(3) Among the existing general OOD methods, EIIL exhibits better performance than other baselines, due to their capability to infer proper environments from the data. Besides, the performances of EIIL also suggest the advantages of inferring environments at representation spaces as opposed to raw feature spaces for TSF's high-dimensional input. These observed advantages align with the considerations made in \ourmethod.

\begin{table*}
\centering
\resizebox{\textwidth}{!}{%
\begin{tabular}{ccc|cc|cc|cc|cc}
\hline
\multicolumn{3}{c|}{Dataset}                                                                                                                                      & \multicolumn{2}{c|}{Exchange}   & \multicolumn{2}{c|}{ILI}        & \multicolumn{2}{c|}{ETTh1}      & \multicolumn{2}{c}{ETTh2}       \\ \hline
\multicolumn{1}{c|}{Type}                                                                                & \multicolumn{1}{c|}{Env. Known?}          & Method     & MSE            & MAE            & MSE            & MAE            & MSE            & MAE            & MSE            & MAE            \\ \hline
\multicolumn{1}{c|}{Base}                                                                                & \multicolumn{1}{c|}{No}                   & ERM        & 0.782          & 0.710          & 3.974         & 1.722         & 0.965          & 0.892         & 3.070          & 1.628          \\ \hline
\multicolumn{1}{c|}{\multirow{7}{*}{\begin{tabular}[c]{@{}c@{}}General \\ OOD\\(Invariant \\Learning)\end{tabular}}}    & \multicolumn{1}{c|}{\multirow{5}{*}{Yes}} & GroupDRO   & 0.781          & 0.715          & 3.721          & 1.888          & 0.880          & 0.863          & 3.192          & 1.647          \\
\multicolumn{1}{c|}{}                                                                                    & \multicolumn{1}{c|}{}                     & IRM        & 0.716          & 0.688          & 3.608          & 1.732          & 0.495          & 0.646          & 2.910          & 1.581          \\
\multicolumn{1}{c|}{}                                                                                    & \multicolumn{1}{c|}{}                     & VREx       & 0.781          & 0.715          & 3.671          & 1.875          & 0.874          & 0.859          & 3.238          & 1.662          \\
\multicolumn{1}{c|}{}                                                                                    & \multicolumn{1}{c|}{}                     & SD         & 0.782          & 0.716          & 3.674          & 1.677          & 0.891          & 0.870          & 3.246          & 1.664          \\
\multicolumn{1}{c|}{}                                                                                    & \multicolumn{1}{c|}{}                     & IB-ERM     & 0.787          & 0.719          & 3.673          & 1.677          & 0.883          & 0.865          & 3.209          & 1.654          \\ \cline{2-11} 
\multicolumn{1}{c|}{}                                                                                    & \multicolumn{1}{c|}{\multirow{1}{*}{No}}  & EIIL       & 0.540          & 0.630          & 3.251          & 1.648          & 0.673          & 0.783          & 1.252          & 1.013       \\ \hline
\multicolumn{1}{c|}{\multirow{2}{*}{\begin{tabular}[c]{@{}c@{}}Temporal\\ Shifts\end{tabular}}} & \multicolumn{1}{c|}{\multirow{2}{*}{NA}}  & RevIN      & 0.169          & 0.296          & 1.350          & 0.867          & 0.108          & 0.248          & 0.236          & 0.387          \\
\multicolumn{1}{c|}{}                                                                                    & \multicolumn{1}{c|}{}                     & NST        & 0.151          & 0.281          & 1.351          & 0.871          & 0.118          & 0.260          & 0.258          & 0.406        \\ \hline
\multicolumn{1}{c|}{\multirow{2}{*}{Hybrid}}                                                             & \multicolumn{1}{c|}{Yes}                  & IRM+RevIN  & 0.160          & 0.291          & 1.328          & 0.863          & 0.105          & 0.244          & 0.234          & 0.381          \\
\multicolumn{1}{c|}{}                                                                                    & \multicolumn{1}{c|}{No}                   & EIIL+RevIN & 0.170          & 0.309          & 1.205          & 0.820          & 0.097          & 0.241          & 0.343          & 0.483          \\ \hline
\multicolumn{1}{c|}{Ours}                                                                                & \multicolumn{1}{c|}{No}                   & FOIL       & \textbf{0.136} & \textbf{0.274} & \textbf{1.047} & \textbf{0.768} & \textbf{0.088} & \textbf{0.224} & \textbf{0.210} & \textbf{0.358} \\ \hline
\multicolumn{3}{c|}{Improvement(\%)}                                                                                                                               & +9.93         & +2.50         & +12.94        & +5.00        & +9.27         & +7.05         & +10.26         & +6.04         \\ \hline
\end{tabular}%
}
\caption{Comparison with existing distribution shifts methods across four datasets using Informer backbone. The best results are in \textbf{bold}. NA means not considering environments. Our \ourmethod outperforms all existing distribution shift  methods on all datasets and both metrics.}
\label{tab:OOD_compar}
\end{table*}

\subsection{Ablation Study}
To demonstrate the effectiveness of each module or loss in \ourmethod, we conduct an ablation study that introduces three ablated versions of \ourmethod:
(1) \ourmethod$\backslash${Suf}: remove the surrogate loss in Eq.~\ref{eq:loss_suf} for decomposing \underline{Suf}ficiently predictable $\YSUF$ (2) \ourmethod$\backslash${TEI}: remove the whole \underline{T}ime-series \underline{E}nvironment \underline{I}nference module detailed in Section~\ref{Sec:TEI},i.e. set the number of environment as 1.(3) \ourmethod$\backslash${LP}: removed the \underline{L}abel \underline{P}ropagation approach in $\MTEI$ in Eq.~\ref{eq:LP}. 
All other experiment setups follow Section~\ref{exp:cmp}. The ablation study results are shown in Figure~\ref{fig:ablation}. 

Though \ourmethod outperforms all ablated versions in forecasting accuracy, all designed modules and loss in \ourmethod show individual effectiveness through the ablation study. Specifically, the performance \ourmethod$\backslash${Suf} drops significantly more than other ablated versions, which indicates the necessity of mitigating unobserved covariate issues when applying invariant learning for TSF. Moreover, \ourmethod$\backslash${TEI} consistently outperforms \ourmethod$\backslash${LP} across all datasets, which validates the effectiveness of preserving the temporal adjacency structure for Time Series Forecasting (TSF).
\subsection{Case Study: Analysis of Inferred Environments}
\begin{figure}[h]
\centering
\subfigure[Ablation study of our method and three ablated versions showing the effectiveness of the model design.]{
\begin{minipage}[t]{0.47\linewidth}
\centering
\includegraphics[width=\linewidth]{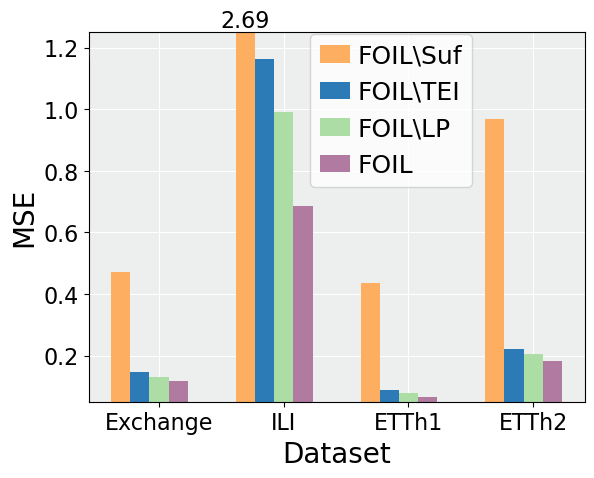}
\label{fig:ablation}
\end{minipage}}\hspace{1mm}
\subfigure[Analysis of two inferred environments on ILI showing significant differences in component weights.]{
\begin{minipage}[t]{0.47\linewidth}
\centering
\includegraphics[width=\linewidth]{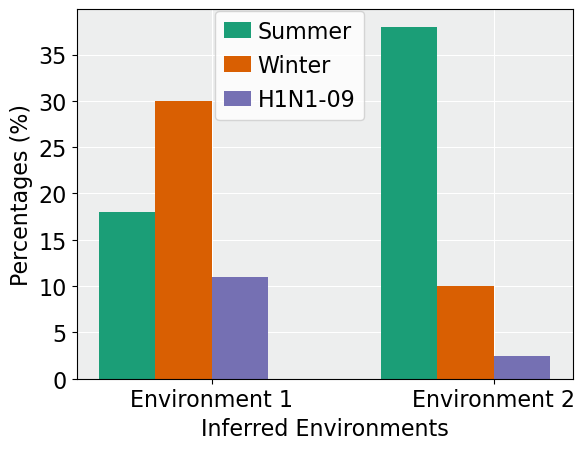}
\label{fig:showcase}
\end{minipage}}
\centering
\caption{Results of analytical experiments.}
\label{fig:combined}
\end{figure}

To justify the reasonableness of the environments inferred by \ourmethod, we conduct a case study on the ILI dataset by demonstrating the contribution disparities among three major components (Summer, i.e., June to August annually; Winter, i.e., December to February annually; and the H1N1-09 period, i.e., April 2009 to August 2010) when the total number of inferred environments is set to 2. The visualization of contributions from each component is shown in Figure~\ref{fig:showcase}. The visualization results align with public health perspectives in two ways: First, the major components of Environment 1 and 2 are distinguished by Winter and Summer, as influenza is a seasonal disease and typically spreads during the winter and ends before the summer. Second, the H1N1-09 period has more contributions in Environment 1 than 2, which aligns with the fact that the H1N1-09 period and winter flu seasons exhibit similarities. These observations support the ability of \ourmethod to infer meaningful environments in real-world TSF applications.

\section{Additional Related Works}\label{sec:relatedworks}
\subsection{Time Series Forecasting}

Classical TSF models~\cite{statis,arima1,arima2} often face limitations in capturing complex patterns due to their inherent model constraints. Recent advancements in deep learning methods, such as Recurrent Neural Networks (RNN) and Transformer~\cite{RNN,LSTM,attention}, have led to sophisticated deep TSF models including Informer, Reformer, Autoformer, Fedformer, and PatchTST~\cite{informer,reformer,autoformer,fedformer,patchtst}, significantly improving forecasting accuracy.
However, these advanced models primarily rely on ERM with simple IID assumptions. Consequently, they exhibit shortcomings in OOD generalization when faced with potential distribution shifts in TS data.

\subsection{Distribution Shifts in Time-Series Forecasting.}

In addition to the aforementioned TSF methods in handling marginal distribution shifts~\cite{DAIN, revin, Nonstationary, dish,adarnn}, there are some efforts that have tackled OOD challenges in TSF. However, all have certain limitations. For example, DIVERSITY~\cite{diversity1,diversity2} is specifically designed for time series classification and detection tasks. OneNet~\cite{onenet} is tailored for online forecasting scenarios by online ensembling. Pets~\cite{pets} focuses on distribution shifts induced by the specific phenomenon of performativity. This highlights the need for a general OOD method applicable across diverse TSF scenarios and models.

Despite the existing benchmark WOODS~\cite{woods} that evaluates IL methods combined with TSF models with a focus on TS classification tasks, our proposed approach addresses diverse datasets under realistic TSF scenarios, offering different and comprehensive problem formulation, methodology, and evaluations.

\section{Conclusion and Discussion}\label{sec:conclusion}

In this paper, we formally study the fundamental out-of-distribution challenges in time-series forecasting tasks (OOD-TSF).  We identify specific gaps when applying existing invariant learning methods to OOD-TSF, including theoretical violations of sufficiency and invariance assumptions and the empirical absence of environment labels in time-series datasets. To address these challenges, we introduce a model-agnostic framework named \ourmethod, which employs an innovative surrogate loss to alleviate the impact of unobserved variables. \ourmethod features a joint optimization strategy, which learns invariant representations and preserves temporal adjacency structure. Empirical evaluations demonstrate the effectiveness of \ourmethod by consistently improving the performances of different TSF models and outperforming other OOD solutions.

Beyond the scope of \ourmethod, it is important to recognize that invariant learning is not the only solution to enhance OOD generalization in TSF tasks. Alternative approaches or interpretations can require advanced causal analysis, feature selections, or learning dynamic temporal patterns. The using of additional information to enhance the sufficiency of predictions also deserves to be explored. We also emphasize the need for conscientious evaluations on underrepresented subgroups when implementing our approach in real-world scenarios for promoting fairness among subgroups. We expect that future research will delve into these open questions, contributing both theoretically and practically to advance the understanding of OOD-TSF challenges and achieve more reliable TSF models.

\section*{Impact Statement}
Our work introduces a new methodology to improve the out-of-distribution generalization of time-series forecasting models and is applicable across wide
range of domains and real-world applications including sensitive applications in public health, economics, etc.
Therefore, care should be taken in alleviating biases and disparities in dataset
as well as making sure the predictions of model do not pose ethical risks
or lead to inequitable outcomes across various stakeholders relevant to
specific applications our methods are used.
\section*{Acknowledgements}
We thank the anonymous reviewers for their helpful comments. This paper was supported in part by the NSF (Expeditions CCF-1918770, CAREER IIS-2028586, Medium IIS-1955883, Medium IIS-2106961, PIPP CCF-2200269, IIS-2008334, CAREER IIS-2144338), CDC MInD program, Meta faculty gifts, and funds/computing resources from Georgia Tech.





\bibliography{A_MAIN}
\bibliographystyle{icml2024}

\newpage
\appendix
\onecolumn
\section{Algorithm}\label{AP:Al}
\begin{algorithm}
\caption{The training procedure of our \ourmethod.}
\begin{algorithmic}
\STATE {\bfseries Require:} Time-series dataset \( \mathcal{D} = \{(\mathbf{X}_i, \mathbf{Y}_i)\}^N_{i=1} \)
\STATE {\bfseries Ensure:} An optimized predictor \( \rho(\phi(\cdot)): \mathcal{X} \rightarrow \mathcal{Y} \)
\STATE Initialize $\rho(\cdot)$, $\{\rho^{(e)}(\cdot)\}$, $\phi(\cdot)$
\STATE Random assign environment label for each $(\mathbf{X}_{i}, \mathbf{Y}_i)$.
\WHILE{not converged}
    \STATE \textbf{Stage 1: Time-series Invariant Learning:} Update $\phi(\cdot)$, $\rho(\cdot)$ according to Equation~\ref{loss:IL}.
    \STATE \textbf{Stage 2: Time-series Environment Inference:}
    \STATE \quad \textbf{M Step:} Fit models according to Equation~\ref{loss:TEI}, update $\{\rho^{(e)}\}$.
    \STATE \quad \textbf{E Step:} Reallocate environment labels according to Equation~\ref{eq:EP} and Equation~\ref{eq:LP}. 
\ENDWHILE
\STATE \textbf{return} $\rho(\cdot)$ and $\phi(\cdot)$.
\end{algorithmic}
\end{algorithm}

\section{Additional Experimental Details}~\label{exp}
\subsection{Datasets}
We conduct experiments on four real-world datasets, commonly used as benchmark datasets:
\begin{itemize}
    \item \textbf{Exchange} dataset records the daily exchange rates of eight currencies. 
    \item \textbf{ETTh1} and \textbf{ETTh2} datasets record the hourly electricity transformer temperature, comprising two years of data collected from two separate counties in China. They include seven variables. We omitted ETTm1 and ETTm2 as they share the same data source as ETTh1 and ETTh2, but with different sampling frequencies.
    \item \textbf{ILI} dataset collects data on influenza-like illness patients weekly, with eight variables. We mainly follow ~\cite{timesnet} to preprocess data, split datasets into train/validation/test sets and select the target variables. All datasets are preprocessed using the zero-mean normalization method.
\end{itemize}
\subsection{Backbones}
As aforementioned, our proposed \ourmethod is a model-agnostic framework. We select three different types of TSF models as backbones. \textbf{Informer}~\cite{informer} proposes an efficient transformer for long-term TSF. \textbf{Crossformer}~\cite{crossformer} better utilizes cross-dimension dependency, making it more sensitive to spuriouse correlations. \textbf{PatchTST}~\cite{patchtst} employs channel-independent and patching strategies to achieve state-of-the-art performance.
\subsection{Baselines: General OOD Methods}
\begin{itemize}
\item \textbf{Methods with Environment Labels:}
    \textbf{IRM}~\cite{IRM} introduces a penalty to learn invariant predictors across different environments. On the basis of the invariance principle of IRM, \textbf{IB-ERM}~\cite{IB-ERM} incorporates the information bottleneck constraint.
\textbf{VREx}~\cite{VREx} propose a penalty on the variance of training risks between environments as a simple agent of risk extrapolation. \textbf{SD}~\cite{SD} proposes a regularization method aimed at decoupling feature learning dynamics to achieve better OOD generalization.\textbf{GroupDRO}~\cite{GroupDRO}, a regularizer for worst-case group generalization, often considered to have general OOD generalization capabilities.
\item \textbf{Methods without Environment Labels:} 
    \textbf{EIIL}~\cite{EIIL} infers the most informative environments for downstream learning invariant predictors by maximizing the penalty in IRM.
\end{itemize}
We omit AdaRNN~\cite{adarnn} for not being model-agnostic; DIVERSITY~\cite{diversity1,diversity2}, as it's specific to time series classification and detection tasks; and multi-view TSF methods~\cite{camul}, which treat each covariate as one view and inflate the parameter count, leading to unfair comparison.
\subsection{Implementation}
For the backbones, we utilize implementations and hyperparameter settings from the Time Series Library\footnote{\url{https://github.com/thuml/Time-Series-Library}}. For general OOD methods, we employ the implementations and tune hyperparameter suggested by DomainBed\footnote{\url{https://github.com/facebookresearch/DomainBed}}. For TSF methods, we use the implementations and hyperparameter settings from their corresponding papers. We have added an MLP to the end of PatchTST to utilize covariates effectively. For our proposed framework \ourmethod, we also incorporate RevIN like PatchTST to address the issue of non-stationarity. We perform affine transformation on each dimension of the raw covariate through learnable weight variables to better find invariant features and improve out-of-distribution generalization capabilities.

 \end{document}